# Probabilistic Causal Reasoning

Thomas Dean[1] and Keiji Kanazawa
Department of Computer Science
Brown University
Box 1910, Providence, RI 02912

## Abstract
Predicting the future is an essential component of decision making. In most situations, however, there is not enough information to make accurate predictions. In this paper, we develop a theory of causal reasoning for predictive inference under uncertainty. We emphasize a common type of prediction that involves reasoning about *persistence*: whether or not a proposition once made true remains true at some later time. We provide a decision procedure with a polynomial-time algorithm for determining the probability of the possible consequences of a set events and initial conditions. The integration of simple probability theory with temporal projection enables us to circumvent problems in dealing with persistence by nonmonotonic temporal reasoning schemes. The ideas in this paper have been implemented in a prototype system that refines a database of causal rules in the course of applying those rules to construct and carry out plans in a manufacturing domain.

## I. Introduction

We are interested in the design of robust inference systems for generating and executing plans in routine manufacturing situations. We hope to build autonomous agents capable of dealing with a fairly circumscribed set of possibilities in a manner that demonstrates both strategic reasoning (the ability to anticipate and plan for possible futures) and adaptive reasoning (the ability to recognize and react to unanticipated conditions). In this paper, we develop a computational theory for temporal reasoning under uncertainty that is well suited to a wide variety of dynamic domains.

The domains that we are interested in have the following characteristics: (i) things cannot always be predicted accurately in advance, (ii) plans made in anticipation of pending events often have to be amended to suit new information, and (iii) the knowledge and ability to acquire predictive rules is severely limited by the planner's experience. Reasoning in such domains often involves making choices quickly on the basis of incomplete information. Although predictions can be inaccurate, it is often worthwhile for a planner to attempt to predict what conditions are likely to be true in the future and generate plans to deal with them.

Our theory includes (i) a polynomial-time decision procedure for probabilistic inference about temporally-dependent information, (ii) a space and time efficient method for refining probabilistic causal rules.

## II. Causal Theories

In order to explore some of the issues that arise in causal reasoning, we will consider some examples involving a robot foreman that directs activity in a factory. The robot has a plan of action that it is continually executing and revising. Among its tasks is the loading of trucks for clients. If our robot learns that a truck is more likely to leave than it previously believed, then it should consider revising its plans so that this truck will be loaded earlier. If, on the other hand, it predicts that all trucks will be loaded ahead of schedule, then it should take advantage of the opportunity to take care of other tasks which it did not previously consider possible in the available time.

In order to construct and revise its plan of action, the robot makes use of a fairly simple model of the world: a special-purpose theory about the cause-and-effect relationships that govern processes at work in the world (referred to as a *causal theory*). The robot's causal theory consists of two distinct types of rules which we will refer to as *projection rules* and *persistence rules*. We will defer discussion of persistence rules for just a bit.

[1] This work was supported in part by the National Science Foundation under grant IRI-8612644 and by an IBM faculty development award. A version of this paper has been presented at CSCSI-88.

As an example of a projection rule, the robot might have a rule that states that if a client calls in an order, then, with some likelihood, the client's truck will eventually arrive to pick up the order. The consequent prediction, in this case the arrival of a client's truck, is conditioned on two things: an event referred to as the *triggering event*, in this case the client calling in the order, and an enabling condition corresponding to propositions that must be true at the time the triggering event occurs. For example, the rule just mentioned might be conditioned on propositions about the type of items ordered, whether or not the caller has an account with the retailer, or the time of day. The simplest form of a projection rule is $PROJECT(P_1 \wedge P_2 \ldots \wedge P_n, E, R, \kappa)$. This says that $R$ will be true with probability $\kappa$ immediately following the event $E$ given that $P_1$ through $P_n$ are true at the time $E$ occurs. Let $\langle P, t \rangle$ indicate that the fluent $P$ is true at time $t$, and $\langle E, t \rangle$ indicate that an event of type $E$ occurs at time $t$. Restated as a conditional probability, this would be:

$$p(\langle R, t+\epsilon \rangle \mid \langle P_1 \wedge P_2 \ldots \wedge P_n, t \rangle \wedge \langle E, t \rangle) = \kappa$$

In this paper, we will assume for simplicity that $P_1$ through $P_n$ are independent. In [4] we discuss methods by which this restriction can be removed. Projection rules are applied in a purely antecedent fashion (as in a production system) by the inference engine we will be discussing. The objective is to obtain an accurate picture of the future in order to support reasoning about plans [2] [1].

Our approach, as described up to this point, is fairly traditional and might conceivably be handled by some existing approach [13] [7]. What distinguishes our approach from that of other probabilistic reasoning approaches is that we are very much concerned with the role of time and in particular the tendency of certain propositions (often referred to as *fluents* [11]) to change with the passage of time. By adding time as a parameter to our causal rules, we have complicated both the inference task and the knowledge acquisition task. Complications notwithstanding, the capability to reason about change in an uncertain environment remains an important prerequisite to robust performance in most domains. We simply have to be careful to circumscribe a useful and yet tractable set of operations. In our case, we have allowed the computational complexity of the reasoning tasks and the availability and ease of acquisition of the data to dictate the limitations of our inference mechanism.

Our inference system needs to deal with the imprecision of most temporal information. Even if a robot is able to consult a clock in order to verify the exact time of occurrence of an observed event, most information the robot is given is imprecise (*e.g.*, a client states that a truck will pick up an order at *around* noon, or a delivery is scheduled to arrive *sometime* in the next 20 minutes). One of the most important sources of uncertainty involves predicting how long a condition lasts once it becomes true (*i.e.*, how long an observed or predicted fact is likely to *persist*). In most planning systems (*e.g.*, [14]) there is a single (often implicit) default rule of persistence [6] that corresponds more or less to the intuition that a proposition once made true will remain so until something makes it false. The problem with using this rule is that it is necessary to predict a contravening proposition in order to get rid of a lingering or persistent proposition: a feat that often proves difficult in nontrivial domains. If a commuter leaves his newspaper on a train, it is not hard to predict that the paper is not likely to be there the next time he rides on that train; however, it is quite unlikely that he will be able to predict what caused it to be removed or when the removal occurred.

When McDermott first proposed the notion of persistence as a framework for reasoning about change [12], he noted that persistence might be given a probabilistic interpretation. That is exactly what we do here. We replace the single default rule of persistence used in most planning systems with a set of (probabilistic) rules: one or more for each fluent that the system is aware of. Our robot might use a persistence rule to reason about the likelihood that a truck driver will still be waiting at various times following his arrival at the factory. The information derived from applying such a rule might be used to decide which truck to help next or how to cope when a large number of trucks are waiting simultaneously. Each persistence rule has the form $PERSIST(P, \rho)$, where $P$ is a fluent and $\rho$ is a function of time referred to as a *survivor* function [17]. In our implementation, we consider only two types of survivor functions: exponential decay functions and piecewise linear functions. The former are described in Section IV., and the latter, requiring a slightly more complex analysis, are described in [5]. Exponential decay functions are of the form $e^{-\lambda t}$ where $\lambda$ is the constant of decay. Persistence rules referring to exponential decay functions are notated simply $PERSIST(P, \lambda)$. Such functions are used, for example, to indicate that the probability of a truck remaining at the dock decreases by 5% every 15 minutes. The persistence rule $PERSIST(P, \lambda)$ encodes the fact that:

$$p(\langle P,t \rangle \mid \langle P, t-\Delta \rangle) = e^{-\lambda(t-\Delta)}$$

where $\Delta$ is a positive number indicating the length of an interval of time. Exponential decay functions are insensitive to changes in the time of occurrence of events that cause such propositions to become true, and, hence, are easy to handle efficiently.

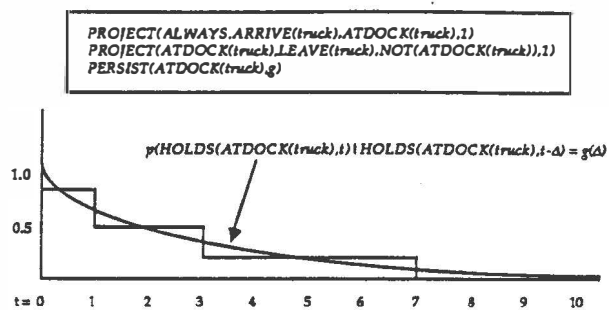

Figure 1: A simple causal theory illustrating the use of survivor functions

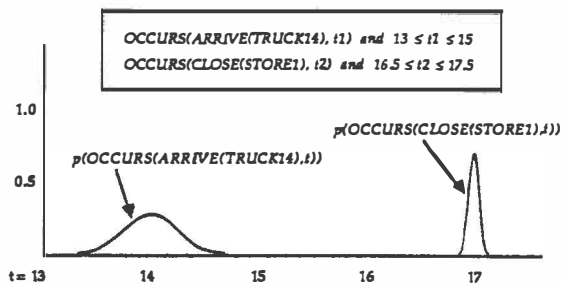

Figure 2: A set of basic facts and their probabilistic interpretation

There are a number of issues that every computational approach to reasoning about causality must deal with. One such issue involves reasoning about dependent causes [13] (*e.g.*, the application of two probabilistic causal rules that have the same consequent effects, both of which appear to apply in a given set of circumstances but whose conditions are correlated). Another issue concerns handling other forms of incompleteness and nonmonotonic inference [9] [3] (*e.g.*, the robot might have a general rule for reasoning about the patience (persistence) of truck drivers waiting to be served and a special rule for how they behave right around lunch time or late in the day). While we agree that these problems are important, we do not claim to have any startling new insights into their solution. There is one area, however, in which our theory does offer some new insights, and that concerns the form of probability functions used in causal rules and how they can be used to efficiently predict the causal consequences.

## III. Probabilistic Projection

In this section, we will try to provide some intuition concerning the process of reasoning about persistence, which we will refer to as *probabilistic projection*. A planner is assumed to maintain a picture of the world changing over time as a consequence of observed and predicted events. This picture is formed by extrapolating from certain observed events (referred to as *basic facts*) on the basis of rules believed to govern objects and agents in a particular domain. These governing rules are collectively referred to as a *causal theory*.

Figure 1 depicts a simple causal theory. Predicates ($ATDOCK$), and constants ($TRUCK14$) are in upper case, while functions ($p$, $g$) and variables ($t$, $truck$) are in lower case. We refer to an instance of a fact (type) being true over some interval of time as a *time token*, or simply *token*. For example, $ARRIVE(TRUCK14)$ denotes a general type of event whereas $\langle ARRIVE(TRUCK14), t \rangle$ denotes a particular instance of $ARRIVE(TRUCK14)$ becoming true. The predicate $ALWAYS$ is timelessly true (*i.e.*, $\forall t \, \langle ALWAYS, t \rangle$). The function $\rho$, a survivor function, describes how certain types of propositions are likely to persist in lieu of further supporting or contravening information.

Figure 2 shows a set of basic facts corresponding to two events assumed in our example to occur with probability 1.0 within the indicated intervals. The system assumes that there is a distribution describing the probability of each event occurring at various times, and uses some default distribution if no distribution is provided.

Evidence concerned with the occurrence of events and the persistence of propositions is combined to obtain a probability function $\pi$ for a proposition $Q$ being true at various times in the future by convolving the density function $f$ for an appropriate triggering event with the survivor function $\rho$ associated with $Q$:

$$\pi(t) = \int_{-\infty}^{t} f(z)\rho(t-z)dz \quad (1)$$

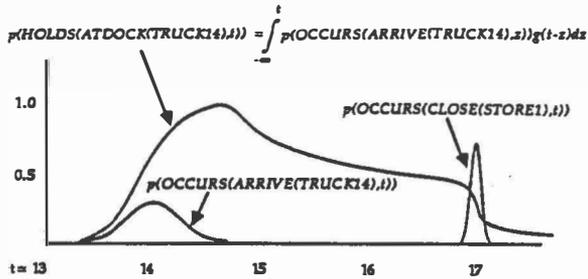

Figure 3: An example of simple probabilistic inference about persistence

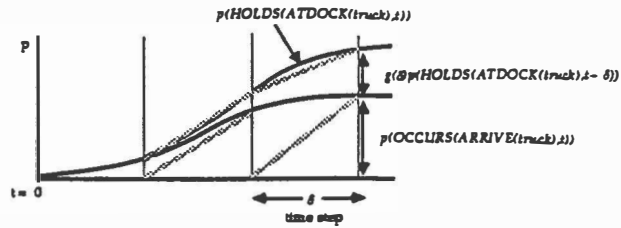

Figure 4: Computing the convolution integral incrementally

Figure 3 illustrates a simple instance of this kind of inference. Note that the range of the resulting probability function is restricted; after the point in time labeled 17, the persistence of $ATDOCK(TRUCK14)$ is said to be *clipped*, and thereafter its probability is represented by another function described below.

All probability computations are performed incrementally in our system. Each token has associated with it a vector which is referred to as its *expectation vector* that records the expected probability that the proposition corresponding to the token's type will be true at various times in the future.

The system updates the expectation vectors every time new propositions are added to the database, and also at regular intervals as time passes. In the update, a single pass sweep forward in time is made through the database. There is, according to the domain and granularity of data, a fixed *time step*, or a quantum by which we partition time. Starting at the "present time," we compute for each proposition its expected probability for the time step according to the causal theory governing that type of proposition, and record it in the expectation vector. We compute the probability for all propositions, before moving on to the next time step. The process is repeated for some finite number of time steps.

For event causation, the update is straightforward; in the simplest cases, it is just a table lookup and copying of the density function into the vector. For the convolution, it is necessary to take steps to avoid computing the convolution integral afresh at each time step. We compute the convolution as a Riemann sum, successively summing over the time axis with a mesh of fixed size (the time step). By using the exponential decay form of survivor functions, it is possible to compute the convolution at a time step by looking only at the value for the last time step, independent of the time at which the proposition of interest became true. All that is required is to multiply the last value by the constant decay rate, and add it to any contribution from the causal distribution for that time step. The process is illustrated graphically in figure 4.

There are many details concerned with indexing and applying projection rules that will not be mentioned in this paper (but see [6]). The details of probabilistic projection using exponential decay functions are described in Section IV.. Our update algorithm is polynomial in the product of the number of causal rules, the size of the set of basic facts, and the size of the mesh used in approximating the integrals. For many practical situations, performance is closer to linear in the size of the set of basic facts.

The convolution equation can be easily extended to handle the case of clipping. We add to (1) a term, the function $g$, corresponding to the distribution of an event which clips the state of a fact being true.

$$\pi(t) = \int_{-\infty}^{t} f(z)e^{-\lambda(t-z)}[1 - \int_{z}^{t} g(w)dw]dz \quad (2)$$

The cumulative distribution of $g$ defines the degree to which it becomes unlikely that the fact represented by $\pi$ remains true in the world. We see that under certain conditions, (2) describes exactly what we desire. Unfortunately, there will be a tendency for the decay function and $g$ to count the same effects twice. In [4] we address methods by which this problem can be attacked in a different framework.

## IV. The Algorithm

Probabilistic causal theories are composed of two types of rules, projection rules:

$$PROJECT(P_1 \wedge P_2 \ldots \wedge P_n, E, R, \kappa)$$

and persistence rules:

$$PERSIST(Q, \lambda)$$

where $P_1$ through $P_n$, $R$, and $Q$ are all fact types, and $E$ is an event type. We assume independence of fact types so that, if we are interested in the conjunction $P_1 \land P_2 \ldots \land P_n$, we can assume that

$$p(\langle P_1 \land P_2 \ldots \land P_n, t \rangle) = \prod_{i=1}^{n} p(\langle P_i, t \rangle) \quad (3)$$

We define a relation $\prec_C$ on fact types so that $Q \prec_C R$ just in case there exists a rule of the form $PROJECT(P_1 \land P_2 \ldots \land P_n, E, R, \kappa)$ where $P_i = Q$ for some $i$. For any given set of causal rules, the graph $\mathcal{G}_{\prec_C}$ whose vertices correspond to fact types and whose arcs are defined by $\prec_C$ is likely to have cycles; this will be the cause of a small complication that we will have to resolve later. In this paper, we distinguish between fact types corresponding to propositions that hold over intervals and event types corresponding to instantaneous (point) events. For each occurrence (token) of a point event of type $E$, we will need its density function $p(\langle E, t \rangle)$. Probabilistic projection takes as input a set of initial events and their corresponding density functions. Given the restricted format for projection rules, the only additional point events are generated by the system in response to the creation of new instances of fact types. For each token of fact type $P$, we identify a point event of type $E_P$ corresponding to the particular instance of that fact becoming true. In the process of probabilistic projection, we will want to compute the corresponding density function $p(\langle E_P, t \rangle)$. In addition to computing density functions, we will also want to compute the mass functions $p(\langle P, t \rangle)$ for instances of facts.

In order to describe the process of probabilistic projection, we will divide the process into two different stages: *deterministic causal projection* and *probabilistic causal refinement*. The actual algorithms are more integrated to take advantage of various pruning techniques, but this simpler, staged, process is somewhat easier to understand. Deterministic causal projection starts with a set of tokens and a set of projection rules and generates a set of new tokens T by scanning forward in time and applying the rules without regard for the indicated probabilities. This stage can be carried out using any number of simple polynomial algorithms (see [6] [10]) and will not be further detailed here. Probabilistic causal refinement is concerned with computing density and mass functions for tokens generated by deterministic causal projection. In the following, all density and mass functions are approximated by step (i.e., piecewise constant) functions. We represent these functions of time using vectors (e.g., $mass(T)$ denotes the mass function for the token $T$ and $mass_i(T)$ denotes the value of the function at $t = i$). For each fact token $T_P$, we create a corresponding event token $T_{E_P}$ and define a vector $mass(T_P)$. For each event token $T_E$, we define a vector $density(T_E)$. We define an upper bound $\Omega$ on projection and assume that each mass and density vector is of length $\Omega$. Initially, we assume that

$$\forall T \in \mathsf{T}: 1 \leq i \leq \Omega: density_i(T) = 0 \land mass_i(T) = 0$$

Event tokens are supplied by the user in the form

$$\kappa = \int_{\text{est}}^{\text{lst}} p(\langle E, t \rangle) dt$$

where est and lst correspond (respectively) to the earliest and latest start time for the token and $\kappa$ is the probability that the event will occur at all. We assume that the density function for such an event is defined by a Gaussian distribution over the interval from est to lst. For a token $T_E$ corresponding to a user-supplied initial event, it is straightforward to fill in $density(T_E)$. Probabilistic causal refinement is concerned with computing $mass_i(T_P)$ and $density_i(T_{E_P})$ for all fact tokens $T_P$ and all event tokens $T_{E_P}$. We partition the set of tokens T into fact tokens $\mathsf{T_F}$ and event tokens $\mathsf{T_E}$. Probabilistic causal refinement can be defined as follows:

```
Procedure: refine( T)
    for i = 1 to Ω:
        for T ∈ T_E: density-update(T, i);
        for T ∈ T_F: mass-update(T, i);
```

Of course, all of the real work is done by density-update and mass-update. Each token has associated with it a specific *derivation* that is used in computing its mass or density. For a token $T_{E_R}$, this derivation corresponds to a rule of the form

$$PROJECT(P_1 \land P_2 \ldots \land P_n, E, R, \kappa)$$

and a set of antecedent tokens $\{T_E, T_{P_1}, T_{P_2} \ldots T_{P_n}\}$ used to instantiate the rule and generate the consequent token $T_R$. Given that

$$p(\langle E_R, t \rangle = \kappa * p(\langle E, t \rangle) * p(\langle P_1 \land P_2 \ldots \land P_n, t \rangle)$$

and, assuming independence (3), we have

```
Procedure: density-update(T_{E_R}, i)
    density_i(T_{E_R}) ←
        κ * density_i(T_E) * ∏_{j=1}^{n} mass_i(T_{P_j})
```

There is one problem with this formulation: it relies on all the mass and density functions for the antecedent conditions being already computed for the instant $i$. In the present algorithm, refine takes no care in ordering the tokens in T. There are a number of ways of ensuring that the updates are performed in the correct order. The easiest is to partially order T according to $\prec_C$ and insist that $\mathcal{G}_{\prec_C}$ be acyclic, but this would preclude the use of most interesting causal theories. A more realistic method is to partition T with respect to an instant $i$ into those tokens that are *open* and those that are *closed*. Deterministic causal projection defines an earliest start time (est) for each token; for event tokens a latest start time (lst) is specified. An event token is open throughout the interval est to lst and closed otherwise. For fact tokens, we modify probabilistic causal refinement so that it closes a fact token $T_P$ as soon as $mass_i(T_P)$ drops below a fixed threshold. A fact token is open from its est until it is closed. All we require then is that for any $i$ the set of tokens that are open define an acyclic causal dependency graph using $\prec_C$. This restriction still allows for a wide range of causal theories. To get refine to do the right thing, we would have to apply refine only to open tokens and either sort the tokens using $\prec_C$, or (as is actually done) define refine so that if, in the course of updating a consequent token, refine finds an antecedent token that hasn't yet been updated, it applies itself recursively.

The derivation of a token $T_P$ corresponds to a rule of the form $PERSIST(P, \lambda)$ where $\lambda$ is the constant of decay for the fact type $P$, and an event token $T_{E_P}$. The procedure mass-update is a bit more difficult to define than density-update since it depends upon the type of decay functions used in persistence rules. In the case of exponential decay functions, the operation of density-update is reasonably straightforward.

Recall the basic combination rule for probabilistic projection:

$$\pi(t) = \int_{-\infty}^{t} f(x)\rho(t-x)dx$$

and suppose that $\rho$ is of the form $e^{-\lambda z}$ where $\lambda$ is some constant of decay, and that $f$ can be approximated by a step function as in

$$f(x) = \begin{cases} C_1 & s_0 \leq x < s_1 \\ \ldots \\ C_n & s_{n-1} \leq x < s_n \end{cases}$$

We will take advantage of the fact that

$$\int_{s_j}^{s_k} f(x)dx = \sum_{i=j}^{k-1} \int_{s_i}^{s_{i+1}} f(x)dx$$

and

$$\rho(s_{k+1} - x) = e^{-\lambda \delta} \rho(s_k - x)$$

where $\delta = s_{k+1} - s_k$.

Making appropriate substitutions, we have

$$\begin{aligned}
\pi(s_{k+1}) &= \sum_{i=j}^{k} \int_{s_i}^{s_{i+1}} f(x)\rho(s_{k+1} - x)dx \\
&= \sum_{i=j}^{k-1} \int_{s_i}^{s_{i+1}} f(x)\rho(s_{k+1} - x)dx \\
&\quad + \int_{s_k}^{s_{k+1}} f(x)\rho(s_{k+1} - x)dx \\
&= e^{-\lambda \delta} \sum_{i=j}^{k-1} \int_{s_i}^{s_{i+1}} f(x)\rho(s_k - x)dx \\
&\quad + \int_{s_k}^{s_{k+1}} f(x)\rho(s_{k+1} - x)dx \\
&= e^{-\lambda \delta} \pi(s_k) + \int_{s_k}^{s_{k+1}} f(x)\rho(s_{k+1} - x)dx
\end{aligned}$$

It should be clear that updates depending upon such simple survivor functions can be performed quite quickly. Integration is approximated using Riemann sums with a mesh of fixed size (roughly) corresponding to $\delta$. We define the procedure mass-update as follows:

Procedure: mass-update($T_P, i$)
   $mass_i(T_P) \leftarrow$
      $e^{\lambda_P \delta} mass_{i-1}(T_P) + density_i(T_{E_P})$

The actual algorithms are complicated somewhat by the fact that the choice of mesh size may not coincide precisely with the steps in the step functions approximating survivor functions and distributions. We compensate for this by using a somewhat finer mesh in the update algorithms. The fact that we employ a fixed mesh size still causes small errors in the accuracy of the resulting mass and density functions, but these errors can be controlled. We have tried to make a reasonable tradeoff, taking into account that the finer the mesh the larger the mass and density vectors. Given that the step functions used for encoding survivor functions and distributions are only approximations, there is a point past which employing a finer mesh affords no additional information. We have found that a mesh size of half the smallest step in any step function works quite well in practice.

## V. Acquiring Rules

Statistical methods have not seen particularly wide application in AI. This is largely due to problems concerning the availability of the data necessary to employ such methods. Data provided from experts has been labeled as unreliable, and the use of priors in Bayesian inference has been much maligned. An alternative to expert judgements and estimating priors, is to integrate the data acquisition process into your system: have it gather its own data. In such a scheme, all predictions made by the system are conditioned only upon what the system has directly observed. Of course, this is unrealistic in many cases (*e.g.*, diagnostic systems whose decisions could impact on the health or safety of humans). In the industrial automation applications considered in this paper, however, not only is it practical, but it appears to be crucial if we are to build systems capable of adapting to new situations.

In this section, we describe a system for continually refining a database of probabilistic causal rules in the course of routine planning and execution. Given the focus of this paper, we will concern ourselves exclusively with the acquisition (or refinement) of persistence rules. Our warehouse planner keeps track of how long trucks stay around and uses this information to construct survivor functions for various classes of trucks. The system must be told which quantities it is to track and how to distinguish different classes of trucks, but given that, the rules it acquires are demonstrably useful and statistically valid.

The survivor function for a given class of trucks is computed from a set of data points corresponding to instances of trucks observed arriving and then observed leaving without being loaded. It should be clear that, in general, a collection of data points will not define a survivor function uniquely. There are many ways in which to derive a reasonable approximation for such a function. For example, we might employ some form of curve fitting based on an expected type of function and the sample data. While such methods may yield more accurate approximations in some cases, for our application, there are simpler and more efficient methods. With both of the simple classes of functions we have considered, the exponential decay and the linear decay functions, computing, respectively, the persistence parameter ($\lambda$) and the slope is trivial. In the case of an exponential decay, we use the mean as the half-life of the function.

We can now sketch the simple algorithm utilized in our system. As noted, we need to collect data for each class of interest. The data for each class is collected in a data structure along with various intermediate quantities used by the update algorithm (*e.g.*, since the algorithm calls for the arithmetic mean of the data points it is convenient to incrementally compute the sum of the elements of the collection). The *class* data type has the following accessor functions associated with it ($c$ is an instance of *class*):

- type($c$): the type of the associated survivor function: linear or exponential
- lambda($c$): the rate or slope
- insts($c$): the number of data points in the collection
- sum($c$): the sum of the items in the collection

Assuming that $c$ is an instance of *class* and $p$ is a new data point, the acquisition algorithm can be described as follows:

```
Procedure: acquire(c, p)
   insts(c) ← insts(c) + 1;
   sum(c)   ← sum(c) + p;
   lambda(c) ← rate(c, sum(c)/insts(c));
```

The function rate depends on the type of survivor function used:

```
Function: rate(c, μ)
   if μ = 0
      then +∞
      else if type(c) = linear
              then 0.5 / μ
              else if type(c) = exponential
                      then (ln 2) / μ
```

Although we have tested our approach extensively in simulations and have found the acquired persistence data to converge very rapidly to the correct values, we do not claim that the above methods have any wider application. The simplicity of the algorithm and its incremental nature are attractive, but the most compelling reason for using it is that the algorithm works well in practice. Probabilistic projection does not rely upon a particular method for coming up with persistence rules. As an alternative, the data might be integrated off line, using more complex (and possibly more accurate) methods.

It should be noted that our system is given the general form of the rules it is to refine. It cannot, on the basis of observing a large set of trucks, infer that trucks from one company are more impatient than those from another company, and then proceed to create two new persistence rules where before there was only one. The general problem of generating

79

causal rules from experience is very difficult. We are currently exploring methods for distinguishing different classes of trucks based on statistical clustering techniques ([8] [15]). Using such methods, it appears to be relatively straightforward to determine that a given data set corresponds to more than one class, and even to suggest candidate survivor functions for the different classes. However, figuring out how to distinguish between the classes in order to apply the different survivor functions is considerably harder.

## VI. Conclusions

In this paper, we have sketched a theory of reasoning about change that extends previous theories [12] [16]. In particular, we have shown how *persistence* can be modeled in probabilistic terms. Probabilistic projection is a special case of reasoning about continuously changing quantities involving partial orders and other sorts of incomplete information, and as such it represents an intractable problem. We have tried to identify a tractable core in the inferences performed by probabilistic projection.

In [5], we describe a planning system capable of continually refining its causal rules. The system makes predictions, observes whether or not those predictions come to pass, and modifies its rules accordingly. It is capable of routine data acquisition and updates its probabilitistic rules in the course of everyday operation. Initial experiments with the prototype system have been very encouraging. We believe that the inferential and causal rule refinement capabilities designed into our system are essential for robots to perform robustly in routine manufacturing situations. We hope that our current investigations will yield a new view of strategic planning and decision making under uncertainty based on the idea of continuous probabilistic projection.